\documentclass{article}

\usepackage[final]{nips_2018}

\usepackage[utf8]{inputenc} 
\usepackage[T1]{fontenc}    
\usepackage{hyperref}       
\usepackage{url}            
\usepackage{booktabs}       
\usepackage{amsfonts}       
\usepackage{nicefrac}       
\usepackage{microtype}      

\usepackage{amsmath,amsfonts,bm}

\def\eqref#1{equation~\ref{#1}}

\def\1{\bm{1}}

\def\vp{{\bm{p}}}

\def\vx{{\bm{x}}}
\def\vy{{\bm{y}}}

\DeclareMathAlphabet{\mathsfit}{\encodingdefault}{\sfdefault}{m}{sl}
\SetMathAlphabet{\mathsfit}{bold}{\encodingdefault}{\sfdefault}{bx}{n}

\def\gG{{\mathcal{G}}}

\def\sG{{\mathbb{G}}}
\def\sH{{\mathbb{H}}}

\def\sS{{\mathbb{S}}}

\usepackage{graphicx}
\usepackage[]{todonotes}

\usepackage{algorithm}
\usepackage[noend]{algpseudocode}
\usepackage{tabularx}
\usepackage{comment}

\usepackage{subfig}
\usepackage{caption}
\usepackage{booktabs}

\newcommand{\G}{\mathcal{G}}

 \newcommand{\Add}[2][]{\todo[inline,linecolor=blue,backgroundcolor=blue!25,bordercolor=blue,#1]{#2}}

\newcommand{\OUR} {{\sc BounceGrad}}

\newcommand{\dtrain}{D_{\it train}}

\usepackage{amsmath,amssymb}

\usepackage{sidecap} 
\usepackage{cleveref}
\usepackage{mathrsfs}
\usepackage{graphicx}
\usepackage{epsfig}
\usepackage{pdfpages}
\usepackage{wrapfig}
\usepackage{lipsum}
\usepackage{times}
\usepackage{setspace}
\usepackage{cancel}
\usepackage{wrapfig}
\usepackage{xspace}
\usepackage{float}
\usepackage{hyperref}

\title{Modular meta-learning in abstract graph networks for combinatorial generalization}

\author{Ferran Alet$^1$, Maria Bauza$^2$,\\ \textbf{Alberto Rodriguez$^2$, Tom\'{a}s Lozano-P\'{e}rez$^1$, and Leslie P. Kaelbling$^1$} \\
$^1$ CSAIL, MIT\\
$^2$ Mechanical Engineering, MIT\\
\texttt{\{alet,bauza,albertor,tlp,lpk\}@mit.edu}
}

\begin{document}

\maketitle

\vspace{-1\baselineskip}
\begin{abstract}
Modular meta-learning is a new framework that generalizes to unseen datasets by combining a small set of neural modules in different ways. In this work we propose \textit{abstract graph networks}: using graphs as abstractions of a system's subparts without a fixed assignment of nodes to system subparts, for which we would need supervision. We combine this idea with modular meta-learning to get a flexible framework with combinatorial generalization to new tasks built in. We then use it to model the pushing of arbitrarily shaped objects from little or no training data. 
\end{abstract}
\vspace{-.5\baselineskip}
\section{Introduction}\vspace{-.5\baselineskip}
Meta-learning, or learning-to-learn, aims at fast generalization. The premise is that by training on a distribution of tasks we can learn a {\em learning algorithm} that, when given a new task, will learn from very little data. Recent progress in meta-learning has been very promising. However, we would like to generalize in broader ways by exploiting inherent structure and compositionality in the domains. 

We use the {\em modular meta-learning} framework~\citep{alet2018modular}, which generalizes by learning a set of neural network {\em modules} that can be composed in different ways to solve a new task, without changing their weights. We generalize to unseen data-sets by combining learned concepts in the domain, exhibiting combinatorial generalization; "making infinite use of finite means"(von Humboldt). 

Combinatorial generalization is also one of the main motivations behind \textit{graph neural networks} (GNNs). Therefore, we suggest that GNNs are an ideal structure to use as a rule for combining modules in modular meta-learning. However,  GNNs are typically only used when there is a clear underlying graph structure in the  domain: sets of entities and relations (such as charged particles or body skeletons) with supervised data for each entity. This limits the domains in which graph networks can be applied. We propose \textit{abstract graph networks}~(AGNs), which use GNNs as an abstraction of a system into its sub-parts without the need to a-priori match each subpart to a concrete entity.  A similar idea underlies finite element methods frequently used in mathematics and engineering.

We show that combining modular meta-learning and AGNs generalizes well by choosing different \textit{node} and \textit{edge} modules, leading to flexible generalization from little data. We apply this method to a new diverse real-world dataset \citep{omnipush} of a robot pushing objects of varying mass distribution and shape on a planar surface. We generalize over both the distribution of objects we trained on and out-of-distribution objects and surfaces from little data. Moreover, with a second model (section \ref{subsec:GEN}) we generalize to new objects with no data, only a diagram of the object. 

\begin{SCfigure}[42][h]
\includegraphics[width=.1\textwidth]{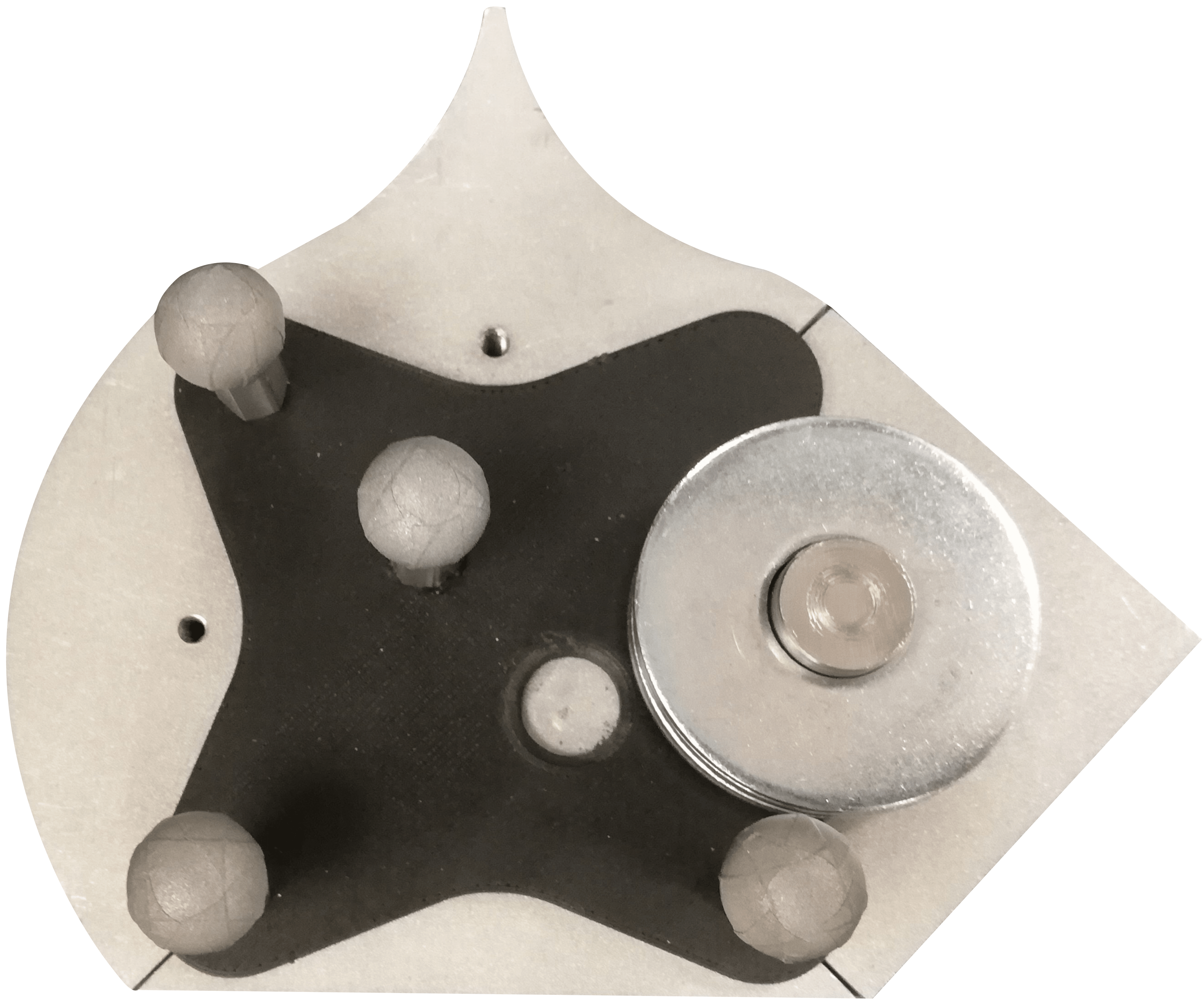}
\caption{Modeled objects are made of the four possible sides. Masses can be attached in the holes. There is an extra hole in the triangle, covered by the wafers.}
\label{fig:sample_object}
\end{SCfigure}

\begin{figure}
    \centering
    \includegraphics[width=\linewidth]{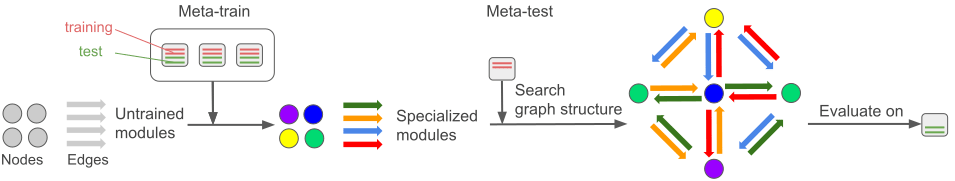}
    \caption{Modular meta-learning with Graph Networks; adapted from \cite{alet2018modular}. 
    }
    \label{fig:parametric-vs-modular-metalearning}\vspace{-1\baselineskip}
\end{figure}

\section{Related Work}
The work closest to ours in the meta-learning literature is by \cite{kopicki2017learning} and \cite{alet2018modular}. However, the former uses a product of experts and needs information about the points of contact. The latter uses modular neural networks with an composition similar to a combination of experts, which we show generalizes less well than our abstract graph neural network model. 

We combine modular meta-learning and AGNs, aiming for combinatorial generalization by composing a set of modules into different structures. This is part of a growing body of work that merges the best aspects of deep learning with structured solution spaces in order to obtain broader generalizations \citep{tenenbaum2011grow,andreas2016neural,meyerson2017beyond,chang2018automatically}. 


By choosing between multiple modules for each edge and nods, our work is related to \cite{kipf2018neural}. However, our meta-learning framework encodes the bias of time-independence and we are able to optimize modules as a group instead of having to select them independently of one another. More related work on GNNs can be found at the beginning of section \ref{subsec:AGN} where we introduce abstract graph networks in its context; additional related work can be found in appendix \ref{sec:extra_related_work}.\vspace{-.5\baselineskip}
\section{Methods}\vspace{-.5\baselineskip}
In this section we describe modular meta-learning, abstract graph networks, graph element networks and how they can be combined.
\subsection{Modular meta-learning}
\textit{Meta-learning}, or learning to learn, can be seen as learning a learning algorithm. More formally, in the context of supervised learning, instead of learning a regressor $f$ with parameters $\Theta$ with the objective that $f(\vx_{test}, \Theta)\approx \vy_{test}$, we aim to learn an algorithm $A$ that takes a small training set $D_{train} = (\vx_{train},\vy_{train})$ and returns a hypothesis $h$ that performs well on the test set:
$$h = A(\dtrain, \Theta)\text{ s.t. } h(\vx_{test}) \approx \vy_{test}\text{; i.e. $A$ minimizes } {\cal L}(A(\dtrain, \Theta)(\vx_{test}), \vy_{test}).
\label{eq:objective}$$
Similar to conventional learning algorithms, we optimize $\Theta$, the parameters of $A$, to perform well. Analogous to the training set, we have a meta-training set of tasks each with its own train and test.

Modular meta-learning~\citep{alet2018modular} learns a set of small neural network modules and generalizes by composing them. In particular, let $m_1,\dots,m_k$ be the set of modules, with parameters $\theta_1,\dots,\theta_k$ and ${\cal S}$ a set of structures that describes how modules are composed. For example, simple compositions can be adding the modules outputs, concatenating them, or using the output of several modules to guide attention over the results of other modules. 

Then $\Theta=\cup \theta_i$, and the algorithm $A$ operates by searching over the set of possible structures ${\cal S}$ to find the one that best fits $\dtrain$, and applies it to $\vx_{test}$. Let $h_{S,\Theta}$ be the function that predicts the output using the modular structure $S$ and parameters $\Theta$. Then:
\begin{equation*}
S^* = {\rm arg}\min_{S \in {\cal S}} {\cal L}(h_{S,\Theta}(\vx_{train}), \vy_{train}) ; \;\;\;\;A(\dtrain, \Theta) = h_{S^*,\Theta}.
\label{eq:bestStructure}
\end{equation*}
\vspace{-\baselineskip}
\subsection{Abstract graph networks}\label{subsec:AGN}
Graph neural networks~\citep{gori2005new,scarselli2009graph,battaglia2018relational} perform computations over a graph, with the aim of incorporating \textit{relational inductive biases}:  assuming the existence of a set of entities and relations between them. In general, GNNs are applied in settings where nodes and edges are \textit{explicit} and match concrete entities of the domain. For instance, nodes can be objects~\citep{chang2016compositional,battaglia2016interaction}, parts of these objects \citep{wang2018nervenet,mrowca2018flexible} or users in a social network. In general, nodes and edges represent explicit entities and relations, and supervised data is required for each component. We propose to use GNNs as an abstraction of a system's sub-parts,without a concrete match between entities and nodes. 


Let $\gG$ be a graph,
we define an encoding function $f_{in}$ that goes from input to initial states of the nodes and edges in $\gG$. We run several steps of message passing \citep{gilmer2017neural}: in each step, nodes read incoming messages, which are a function of their neighbours' hidden states and their corresponding edges, and use their current state and the sum of incoming messages to compute their new state. 
Finally, we define a function $f_{out}$ that maps the nodes final states to the final output: $f_{out}\left(MP^{(T)}\left(f_{in}(x)\right)\right)$; trainable end-to-end. This defines a means of combination for the modules, giving us different (differentiable) losses depending on the module we put in each position.

\subsection{Graph element networks} \label{subsec:GEN}
\textit{Finite element methods}(FEMs) are a widely used tool in engineering to solve boundary-value problems for partial differential equations. They subdivide a complex problem in a domain (such as analyzing stress forces on a bridge or temperature in a building) into a mesh of small elements where the dynamics of the PDE can be approximated. With that inspiration, we propose a subtype of abstract graph network, which we denote \textit{graph element networks (GEN)}. In GENs, we \textit{place} nodes in a region (usually of $\mathbb{R}^2$ or $\mathbb{R}^3$); each having specific coordinates. These nodes induce a \textit{Voronoi diagram} where a node is responsible for all points in space to which it is the closest representative.

As in general AGNs, we have to define a function between input and graph input ($f_{in}$) and between graph output and final output ($f_{out}$) and the connectivity between the nodes. 1) \textbf{$f_{in}$}: a natural choice is to map the input to an initial 'vector field' at a particular position in space. We can give the input to the node responsible for the corresponding Voronoi region. For example, in our pushing example we pick the initial position of the pusher and feed the input (including the said initial position) to its closest node. 2) Connectivity: A simple choice is a regular grid over the region, thus also defining the edges. A more general solution, with many desirable properties, is to use the edges from the Delaunay triangulation of the nodes. The Delaunay triangulation is the graph where two nodes are connected if and only if their corresponding Voronoi regions share an edge. 3) \textbf{$f_{out}$}: a learned integrator function that either sums or averages a transformation over the final states of each module; thus being independent of the order and number of nodes.

In GENs nodes are still abstract (meaning they do not have a 1-to-1 correspondence with a discrete element in the system), but they are \textit{grounded} in a particular subregion. This makes it possible for other data to inform the types of node modules to put in each node. For example, in our experiments with pushing objects, the node types are informed by whether there is a mass, regular surface or empty space in that particular position in space. This information can either completely determine the node type (as in our current experiments) or only partially inform it: for instance we could imagine knowing the shape of an object but not the density of each subpart.

Note that, in contrast with \textit{finite element methods}, the dynamics of the system follow completely unknown equations; giving us a more complicated, but also more flexible, framework. GEN nodes and edges have different types, allowing dynamics to be different in different parts of the space; for example depending on the material properties of that particular region. Moreover, we will also have a similar trade-off between accuracy and computational performance where a bigger graph of smaller Voronoi regions will have higher accuracy at the expense of more computational cost.

When graph structures are fixed and module types are completely determined by an external source (such as a picture of the object), we are doing multitask learning between the meta-train tasks and then transferring to new domains without any need for experimental data; only using the picture of the object. However, in general the nodes may not be known (a material not seen at training) or we may customize the graph to the system to get the best accuracy given fixed computational resources.

\begin{SCfigure}[42][h]
    \includegraphics[width=.6\linewidth]{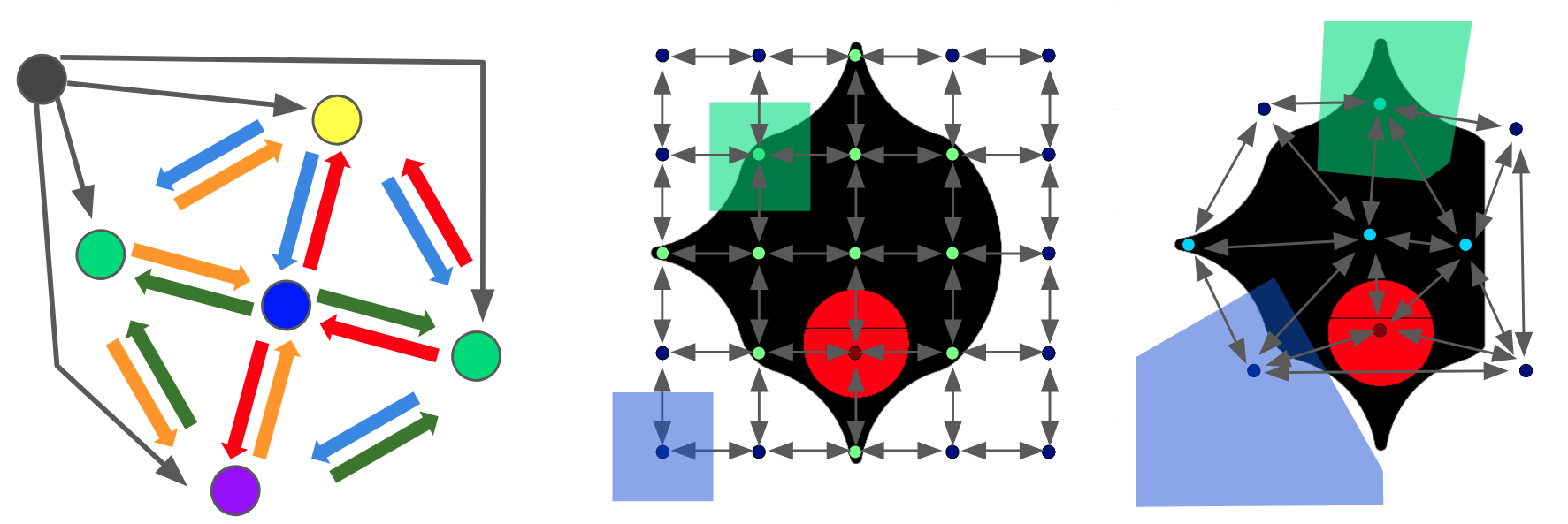}
\caption{\textbf{Left:} AGN wheel graph plus 'pusher' node (details in appendix \ref{sec:wheelnet}). \textbf{Center:} GEN with a 5x5 grid, nodes conditioned on image; highlighted two Voronoi regions. \textbf{Right:} not implemented GEN with Delaunay edges and optimized positions}
\label{fig:sample_object}
\end{SCfigure}
\vspace{-\baselineskip}
\section{Experiments}\vspace{-.5\baselineskip}


The dataset \citep{omnipush}, consists of 250 pushes for each of 250 objects, which were constructed by attaching 4 pieces to a central part with magnets and possibly adding masses, such as the one shown in figure. We also collected data for out-of-distribution objects and a different surface. 

For pushing, we train two structures. 1) a few-shot learning abstract graph network structure which models the object as a wheel graph with bidirectional edges. This structure has many desirable properties: low graph diameter, sparsity and a notion of locality (the position of nodes matters). All details can be found in appendix \ref{sec:wheelnet}. 2) A multitask learning graph element network, where we also receive a top-view image for each object, from which we can extract a model without the need for any data. For the GEN, the graph is a 5 by 5 grid on $[-10cm,10cm]^2$ with the module type of each node is entirely determined by the 'material' at its position (empty, small/big mass, no mass). 


We normalize each dimension of input and output so that the MSE (mean square error) of each coordinate is 1 across all meta-training dataset points. This setting allows us to create a loss that combines angle changes with position changes, which have different orders of magnitude. As a result, the baseline of predicting the current position for the next position (no motion) has MSE 1, and allows to interpret the performance of other algorithms as a percentage of the total variance.

We use algorithm \ref{alg:bouncegrad} to train our system on 200 objects/tasks and test on 50 tasks; within each task we train on 50 points and test on 200. We compare against an object-independent regressor that has a single training set with the full 250 points for 200 objects($50k$ points), and evaluates on the 50 remaining tasks.  Additionally, we compare to modular meta-learning but on an attention-based non-AGN structure used in our previous work.  We find that the method with pooled data performs the worst, outperformed by both modular meta-learning approaches, and that the use of AGNs significantly improves performance over the original modular structure.
See table~\ref{table:results} for all results. 

We also picked 5 objects for which we collected 2500 points, trained a single model for each of them on 2000 points and tested on 500 and obtained a MSE of $0.04$. Previous experiments have shown that the test loss stops decreasing significantly after 2000 points \citep{bauza2017probabilistic}, which leads us to believe this is a good approximation of the Bayes error rate: the irreducible error due to camera and actuation noise, position on the surface, and other factors. Therefore, even though our meta-learned models only use 50 points we are very close to the estimated Bayes error.
\begin{table}[h]
  \centering
  \small
  \tabcolsep=0.11cm{
  \begin{tabular}{lrr}
    \toprule[1pt]
      \textbf{Evaluation}& MSE & distance equivalent\\ 
    \midrule[1pt]
      Predicting no movement & 1.00 & 21.6 mm\\
      \hline
      Estimated Bayes error rate& .04 & 4.3 mm\\
      \hline
      Original distribution (no meta-learning) & .14 & 8.1 mm\\
      Original distribution (\cite{alet2018modular}) & .08 & 6.1 mm \\
      \textbf{Original distribution (meta-learning \& AGN)} & .06 & 5.3 mm\\
      \textbf{Original distribution (image-conditioned \& GEN)} & .05 & 4.7 mm\\
      \hline
      Out of distribution obj.\&surface (no meta-learning) & .51 & 15.4 mm\\ 
      Out of distribution obj.\&surface (\cite{alet2018modular}) & .34 & 12.5 mm\\ 
      Out of distribution obj.\&surface (meta-learning \& AGN) & .29 & 11.6 mm\\ 
      \hline
      Out of distribution objects (no meta-learning)& .34 & 12.7 mm\\
      Out of distribution objects (\cite{alet2018modular}) & .30 & 11.8 mm\\ 
      Out of distribution objects (meta-learning \& AGN) & .25 & 10.8 mm\\ 
      \hline
      Out of distribution surface (no meta-learning) & .21 & 9.8 mm\\ 
      Out of distribution surface (\cite{alet2018modular}) & .09 & 6.6 mm\\ 
      Out of distribution surface (meta-learning \& AGN) & .08 & 5.9 mm\\ 
    \bottomrule[1pt]
  \end{tabular}}
  \caption{Summary of results. Different datasets are separated by horizontal lines.}
  \label{table:results}
  \vspace{-1\baselineskip}
\end{table}

We evaluate on 11 shapes that were not constructed using the four sides and magnets (see \cite{omnipush} for details). We also evaluated the effect of using a different surface (plywood) rather than the one used for the training data. In both cases we observe a predictable drop in performance, but we account for more than $70\%$ of the variance, even when both factors change.

\bibliography{iclr2019_conference}
\bibliographystyle{iclr2019_conference}
\appendix
\section{\OUR\ algorithm for abstract graph networks}

\newcommand{\randomelt}{random\_elt}
\textbf{Notation}
\begin{itemize}
\item $\gG$: graph, with node $n_1,\dots, n_r$ and \textit{directed} edges $e_1,\dots,e_{r'}$.
\item $f_{in}$: encoding function from input to graph initial states.
\item $f_{out}$: decoding function from graph final states to output.
\item $\sG$: set of node modules $g_1,\dots,g_{|\sG|}$, where $g_i$ is a network with weights $\theta_{g_i}$.
\item $\sH$: set of edge modules $h_1,\dots,h_{|\sH|}$, where $h_i$ is a network with weights $\theta_{h_i}$.
\item $S$: a structure, one module per node $m_{n_1},\dots,m_{n_r}$, one module per edge $m_{e_1},\dots,m_{e_{r'}}$. Each $m_{n_i}$ is a pointer to $\sG$ and each $m_{e_j}$ is a pointer to $\sH$.
\item $\mathcal{T}^1,\dots,\mathcal{T}^k$: set of regression tasks, from which we can sample $(x,y)$ pairs.
\item MP$^{(T)}(\gG,S)(x_t)\rightarrow x_{t+1}$: message-passing function applied $T$ times.
\item $\mathcal{L}(y_{target}, y_{pred})$: loss function; in our case $|y_{target}-y_{pred}|^2$.
\item $L$: instantiation of a loss. This includes the actual loss value and infrastructure to backpropagate it.
\item \randomelt$(\sS)$: pick element from set $\sS$ uniformly at random.
\end{itemize}
\begin{algorithm}[h!]
  \caption{Training modular-metalearning on AGNs}
  \small
  \label{alg:bouncegrad}
  \begin{algorithmic}[1]
    \Procedure{InitializeStructure}{$\gG,\sG,\sH$}\Comment{Initialize with random modules}
        \For{$n_i\in\G.nodes$}
            $S.m_{n_i} \gets $\randomelt($\sG$)
        \EndFor
        \For{$e_i\in\G.edges$}
            $S.m_{e_i} \gets $\randomelt($\sH$)
        \EndFor
    \EndProcedure
    \Procedure{ProposeStructure}{$S,\gG,\sG,\sH$}
    \State $P \gets S$
    \If{Bernouilli($1/2$)}
        \State $idx \gets \randomelt(\gG.nodes)$; $P.m_{n_{idx}} \gets $\randomelt($\sG$)
    \Else
        \State $idx \gets \randomelt(\gG.edges)$; $P.m_{e_{idx}} \gets $\randomelt($\sH$)
    \EndIf
        \State \textbf{return} $P$
    \EndProcedure
    \Procedure{Evaluate}{$\gG,S,\mathcal{L},\vx,\vy$}
        \State $\vp \gets f_{out}\left(\text{MP}^{(T)}(\gG,S)(f_{in}(\vx))\right)$ \Comment{Running the AGN with modular structure S}
        \State \textbf{return} $\mathcal{L}(\vy,\vp)$
    \EndProcedure
    \Procedure{\OUR}{$\gG,f_{in},f_{out},\sG,\sH,\mathcal{T}^1,\dots,\mathcal{T}^k$} \Comment{Modules in $\sG,\sH$ start untrained}
    \For{$l\in [1,k]$} 
        \State $S^l \gets$ InitializeStructure$(\gG,\sG,\sH)$
    \EndFor
    \While{not done}
      \State $l \gets \randomelt([1,k])$
      \State $P \gets$ ProposeStructure$(S^l,\gG,\sG,\sH)$
      \State $(\vx,\vy) \gets$ sample$(\mathcal{T}^l)$\Comment{Train data}
      \State $L_{S^l} \gets $Evaluate($\gG,S^l,\mathcal{L},\vx,\vy$)
      \State $L_P \gets $Evaluate($\gG,P,\mathcal{L},\vx,\vy$)
      \State $S^l \gets $SimulatedAnnealing$((S^l,L_{S^l}), (P,L_P))$ \Comment{Choose between $S^l$ and $P$}
      \State $(\vx',\vy') \gets$ sample$(\mathcal{T}^l)$\Comment{Test data}
      \State $L \gets $Evaluate($\gG,S^l,\mathcal{L},\vx',\vy'$)
      \For{$h\in \sH$}
        $\theta_h \gets $GradientDescent($L,\theta_h$)
      \EndFor
      \For{$g\in \sG$}
        $\theta_g \gets $GradientDescent($L,\theta_g$)
      \EndFor
    \EndWhile
    \State \textbf{return} $\sG,\sH$ \Comment{Return specialized modules}
    \EndProcedure
  \end{algorithmic}
  \caption{Alternating optimization between structures and modules on abstract graph networks.}
\end{algorithm}

\section{Extra related work} \label{sec:extra_related_work}
The literature in meta-learning~\citep{schmidhuber1987evolutionary,thrun2012learning,lake2015human} and multi-task learning\citep{torrey2010transfer} is now too extensive to cover entirely, with~\citep{koch2015siamese,vinyals2016matching,duan2017one, mishra2018simple,finn2017model,DBLP:journals/corr/abs-1803-02999} being some prominent examples. The work closest to ours in the meta-learning literature is by \cite{kopicki2017learning} and \cite{alet2018modular}. However, the former uses a product of experts and needs information about the points of contact. The latter uses modular neural networks with an approach similar to a combination of experts, which we show generalizes less well than our graph neural network model. 

We join modular meta-learning \citep{alet2018modular} and abstract graph networks, aiming for combinatorial generalization \citep{humboldt1999language,chomsky2014aspects} by composing a set of modules in different structures. In this sense, we are part of a growing number of works aiming to merge the best of deep learning with structure aiming at broader generalizations \citep{tenenbaum2011grow,andreas2016neural,meyerson2017beyond,xu2017neural,fernando2017pathnet,DBLP:journals/corr/abs-1711-01239, marcus2018deep,pearl2018theoretical,liang2018evolutionary,battaglia2018relational,chang2018automatically,chitnis2018planning,ellis2018search}.

We are part of a growing body of work that merges the best aspects of deep learning with structured solution spaces in order to obtain broader generalizations \citep{tenenbaum2011grow,andreas2016neural,meyerson2017beyond,xu2017neural,fernando2017pathnet,DBLP:journals/corr/abs-1711-01239, marcus2018deep,pearl2018theoretical,liang2018evolutionary,battaglia2018relational,chang2018automatically,chitnis2018planning,ellis2018search}.

Graph neural networks~\citep{gori2005new,scarselli2009graph,gilmer2017neural,battaglia2018relational} perform computations over a graph, with the aim of incorporating \textit{relational inductive biases}:  assuming the existence of a set of entities and relations between them. In general, GNNs are applied in settings where nodes and edges are \textit{explicit} and match concrete entities of the domain. For instance, nodes can be objects~\citep{chang2016compositional,van2018relational,battaglia2016interaction,watters2017visual} or be parts of these objects \citep{wang2018nervenet,SanchezGonzalez2018GraphNA,battaglia2016interaction,mrowca2018flexible,li2018learning}. In general, nodes and edges represent explicit entities and relations, and supervised data is generally required for each component.

The interest on pushing is not new and it has been extensively studied theoretically \citep{mason1986mechanics,peshkin1988motion,lynch1992manipulation,liu2011pushing,behrens2013robotic,zhou2017pushing}, with pure learning methods \citep{finn2016unsupervised,agrawal2016learning,Pinto2016TheCR,bauza2017probabilistic,alet2018modular} and by leveraging theory for machine learning methods \citep{hogan2017reactive,ajayIROS2018,byravan2017se3}. 

\section{All details about our \textit{Abstract Graph Network} for pushing} \label{sec:wheelnet}
We can describe the dynamics of the object as a function of the dynamics of its parts. To do so, we model the object as a graph, and more concretely, a wheel graph. A wheel graph has the property of having a very low diameter (2) yet, in contrast to a star graph, also has a notion of 'locality'. Moreover, all exterior nodes are computationally equivalent, which will make it easier to share modules between them.

The pusher is modeled via a single node. This node will interact with all the other nodes via the same 'edge module'. Since we would like the pusher to only interact with a very small part of the object we will use a (learned) attention mechanism, ensuring that the module node sends a stronger signal to a single part of the object. This is similar to \cite{velickovic2017graph} using attention in the outgoing direction instead of incoming edges.

Our module composition is to build a graph neural network with the shape of a wheel graph. Therefore, for a wheel of $N$ exterior nodes we will need $4N$ 'edge modules': $N$ going clockwise, $N$ counter-clockwise, $N$ to the center and $N$ out from the center. We then have $N+1$ node modules ($N$ exterior, 1 central). Filling these slots with different modules will give us different structures, which in turn will lead to different predictions. The pusher node and edge modules are constant across datasets. 

As we mentioned earlier, all the exterior nodes are symmetric; which doesn't allow us to have a sense of direction. To break this symmetry set a custom 'code' to each node's hidden state; modules can read, but not write, to these parts of the hidden state. The code for the $i$-th exterior node to $[\cos \frac{2\pi i}{N}, \sin \frac{2\pi i}{N}, 0,0,0,0,0]$. The central node is initialized with $[0,0,1,0,0,0,0,0]$ and the pusher with $[0,0,0,1,X_x,X_y,X_\theta]$, initially being the only node containing the input.

We run 5 steps of message passing; enough for the information to go from the pusher to node A(1 step), then node B (2 steps), and finally to C(2 steps) with the goal of capturing global dynamics. After these 5 steps we pick the last 3 positions of the each exterior node and multiply them by the cosine of the corresponding angle, we pick the -6:-3 positions and multiply them by the sine of the corresponding angle and average all these vectors into a 3-dimensional output. Note that, since the integral of sine and cosine is 0, the Graph Neural Network is forced to learn to place different states in different nodes.

Finally, note that we fix the angles, not the concrete positions. Since these angles can be defined for all shapes; this abstraction naturally extends to all 2d shapes. Moreover, we expect (and experimentally show) that modules can be reused to play similar local and global effects on very different shapes. We conjecture that the same architecture could model rich object-object interactions: this could be achieved by having multiple wheels and edges with attention created between nearby objects at each timestep; similar to \cite{chang2016compositional}.

\end{document}